\documentclass{article}

\usepackage{microtype}
\usepackage{graphicx}
\usepackage{subcaption}
\usepackage{booktabs} 

\usepackage{hyperref}

\usepackage[algo2e,ruled,vlined]{algorithm2e}

\usepackage[accepted]{icml2024}

\usepackage{amsmath}
\usepackage{amssymb}
\usepackage{mathtools}
\usepackage{amsthm}
\usepackage{physics}

\usepackage[capitalize,noabbrev]{cleveref}

\usepackage[textsize=tiny]{todonotes}

\usepackage[export]{adjustbox}
\usepackage{orcidlink}
\usepackage[misc]{ifsym}

\usepackage{xspace}
\newcommand{\ifor}{\textsc{IFOR}\xspace}
\newcommand{\oifor}{o\textsc{IFOR}\xspace}
\newcommand{\asdifor}{asd\textsc{IFOR}\xspace}
\newcommand{\hst}{\textsc{HST}\xspace}
\newcommand{\rrcf}{\textsc{RRCF}\xspace}
\newcommand{\loda}{\textsc{LODA}\xspace}
\newcommand{\onlineiforest}{\textsc{Online}-\textsc{iForest}\xspace}
\newcommand{\onlineitree}{\textsc{Online}-\textsc{iTree}\xspace}
\newcommand{\onlineitrees}{\textsc{Online}-\textsc{iTree}s\xspace}
\DeclareRobustCommand{\vect}[1]{
  \ifcat#1\relax
    \boldsymbol{#1}
  \else
    \vb*{#1}
\fi}

\icmltitlerunning{Online Isolation Forest}

\begin{document}
    \twocolumn[
        \icmltitle{Online Isolation Forest}
        
        
        
        \icmlsetsymbol{equal}{*}
        \begin{icmlauthorlist}
            \icmlauthor{Filippo Leveni}{polimi}
            \hspace{-0.17cm}\textsuperscript{(\Letter)} \orcidlink{0009-0007-7745-5686}\hspace{0.15cm}
            \icmlauthor{Guilherme Weigert Cassales}{waikato}
            \hspace{-0.15cm}\orcidlink{0000-0003-4029-2047}\hspace{0.15cm}
            \icmlauthor{Bernhard Pfahringer}{waikato}
            \hspace{-0.15cm}\orcidlink{0000-0002-3732-5787}\hspace{0.15cm}
            \icmlauthor{Albert Bifet}{waikato}
            \hspace{-0.15cm}\orcidlink{0000-0002-8339-7773}\hspace{0.15cm}
            \icmlauthor{Giacomo Boracchi}{polimi}
            \hspace{-0.15cm}\orcidlink{0000-0002-1650-3054}
        \end{icmlauthorlist}
        
        \icmlaffiliation{polimi}{Dipartimento di Elettronica, Informazione e Bioingegneria, Politecnico di Milano, Milan, Italy}
        \icmlaffiliation{waikato}{Artificial Intelligence Institute, University of Waikato, Hamilton, New Zealand}
        
        \icmlcorrespondingauthor{Filippo Leveni}{filippo.leveni@polimi.it}
        
        \icmlkeywords{Anomaly Detection, Online Data, Data Streams, Isolation Forest}
        
        \vskip 0.3in
    ]
    
    
    
    \printAffiliationsAndNotice{}  
    
    \begin{abstract}
        The anomaly detection literature is abundant with offline methods, which require repeated access to data in memory, and impose impractical assumptions when applied to a streaming context.
        Existing online anomaly detection methods also generally fail to address these constraints, resorting to periodic retraining to adapt to the online context.
        We propose \onlineiforest, a novel method explicitly designed for streaming conditions that seamlessly tracks the data generating process as it evolves over time.
        Experimental validation on real-world datasets demonstrated that \onlineiforest is on par with online alternatives and closely rivals state-of-the-art offline anomaly detection techniques that undergo periodic retraining.
        Notably, \onlineiforest consistently outperforms all competitors in terms of efficiency, making it a promising solution in applications where fast identification of anomalies is of primary importance such as cybersecurity, fraud and fault detection.
    \end{abstract}

    \begin{figure*}[t]
        \begin{subfigure}[t]{0.2\textwidth}
            \begin{center}
            \centerline{\includegraphics[width=\columnwidth]{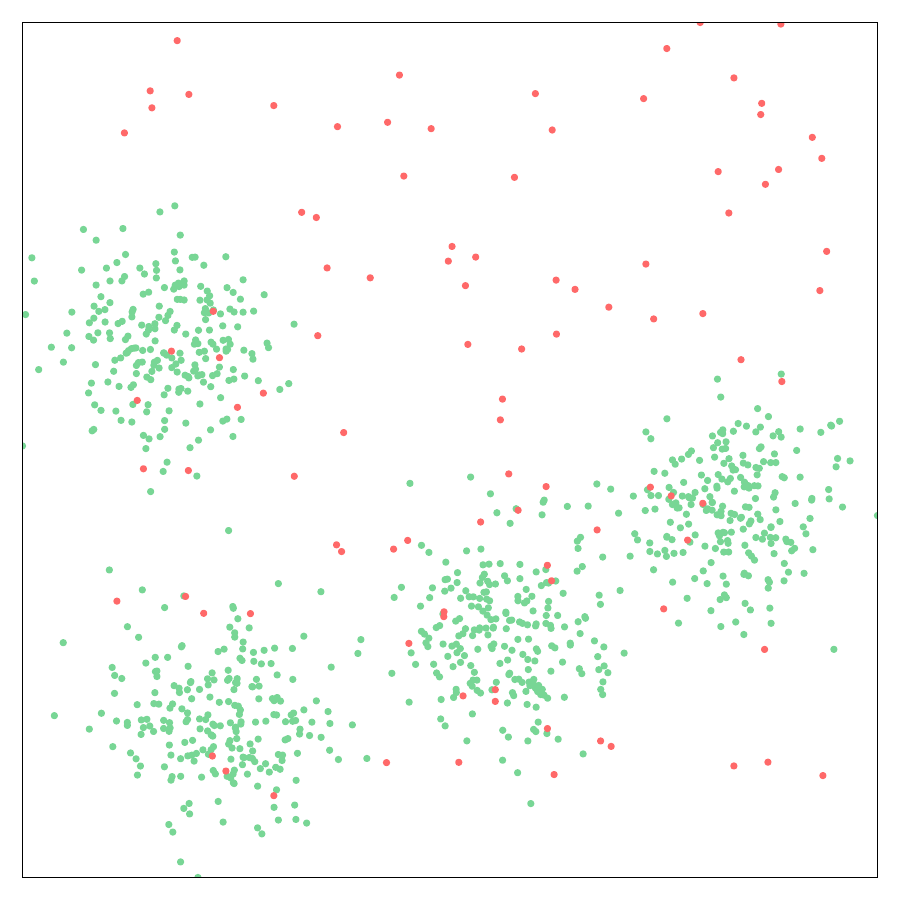}}
            \begin{minipage}{1.25\textwidth}
                \vspace*{-0.225\textwidth}
                \caption{Data stream $\vect{x}_1, \dots, \vect{x}_t \in \mathbb{R}^d$.}
                \label{fig:data}
            \end{minipage}
            \end{center}
        \end{subfigure}
        \hfill
        \begin{subfigure}[t]{0.645\textwidth}
            \begin{center}
            \centerline{\includegraphics[width=0.3125\columnwidth]{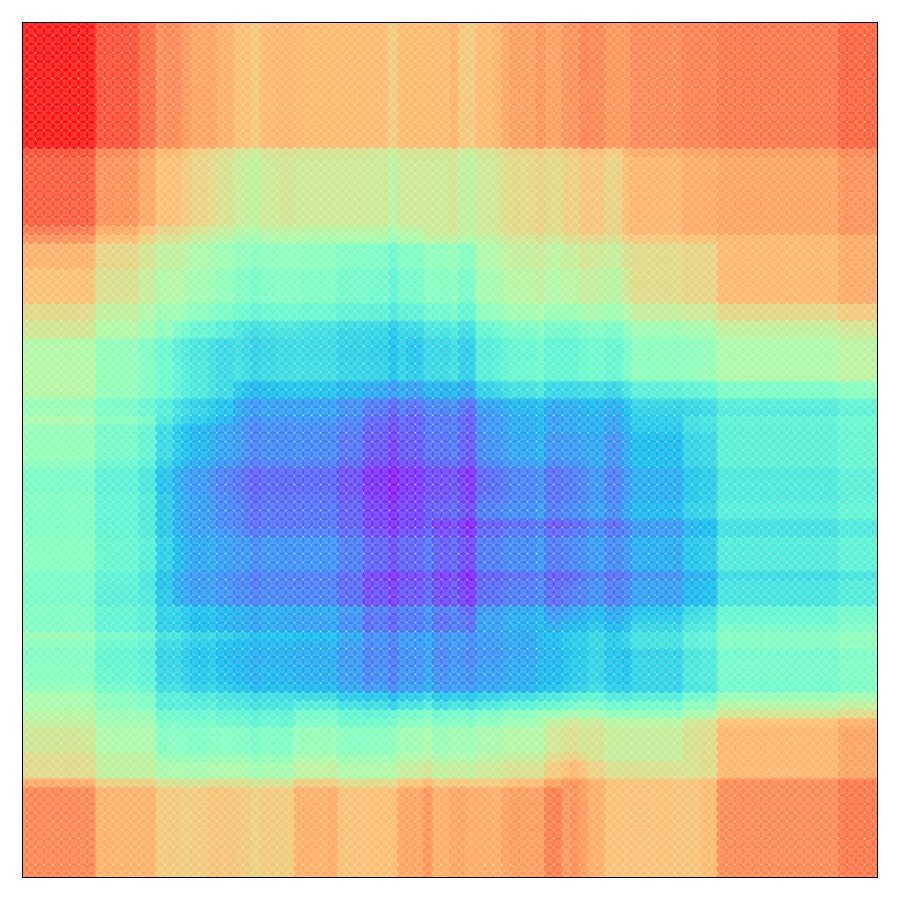} \; \includegraphics[width=0.3125\columnwidth]{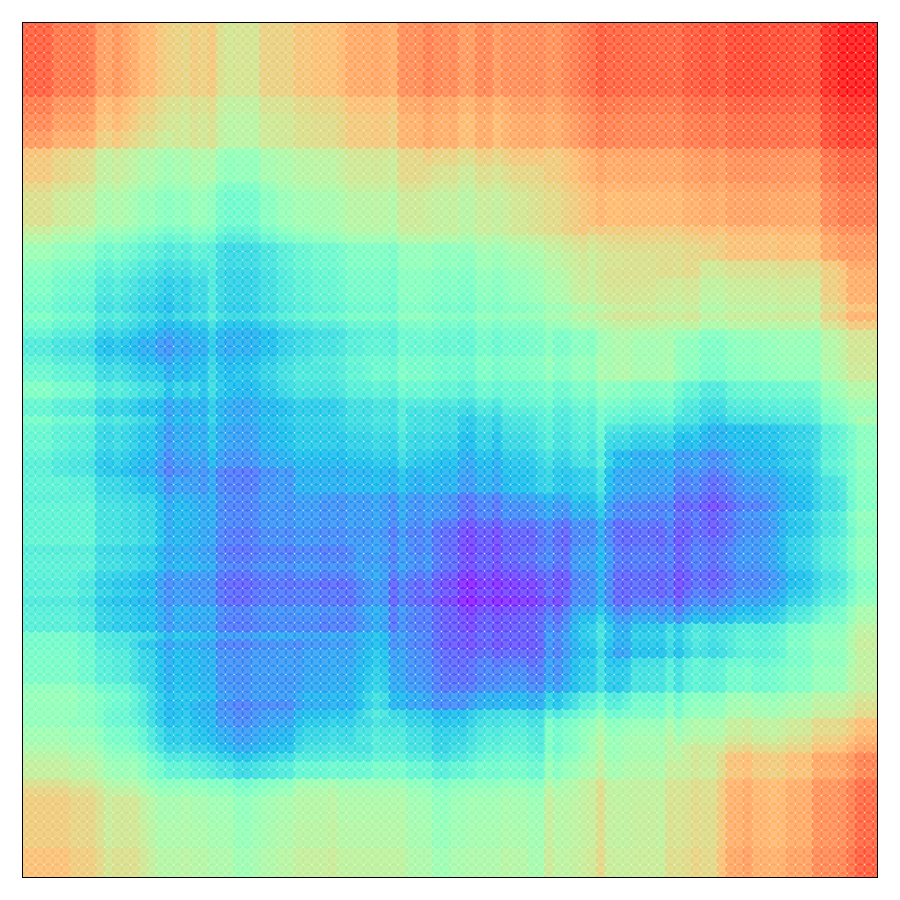} \; \includegraphics[width=0.3125\columnwidth]{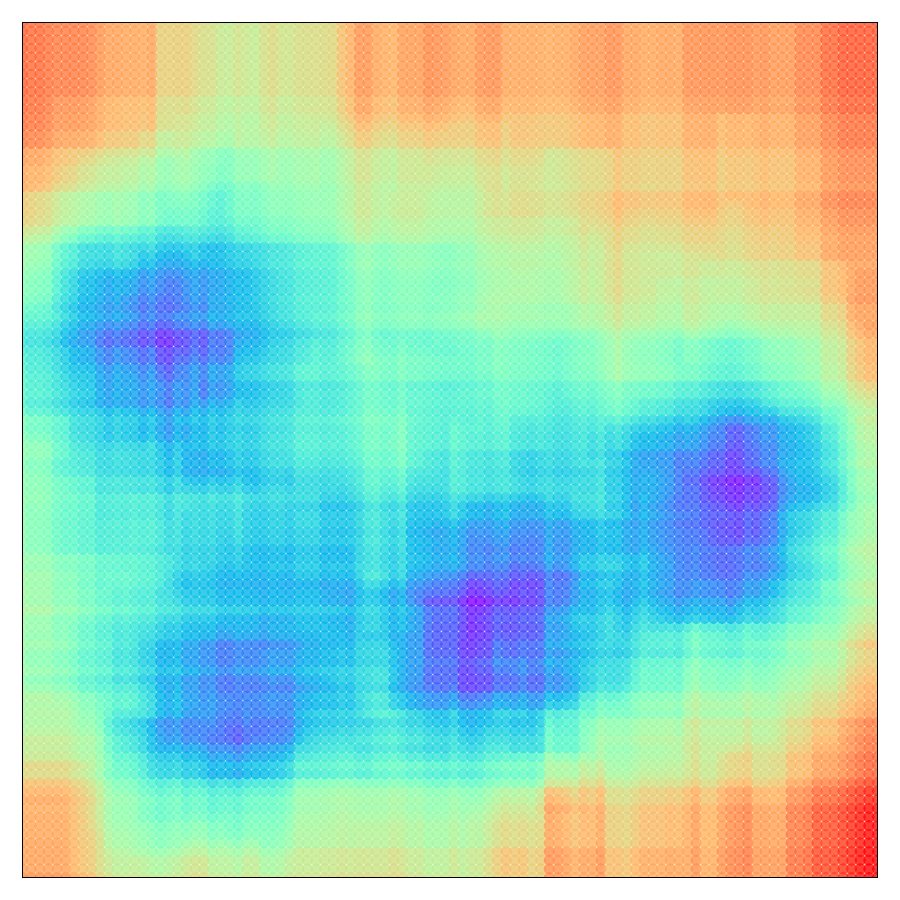} \includegraphics[width=0.045\columnwidth]{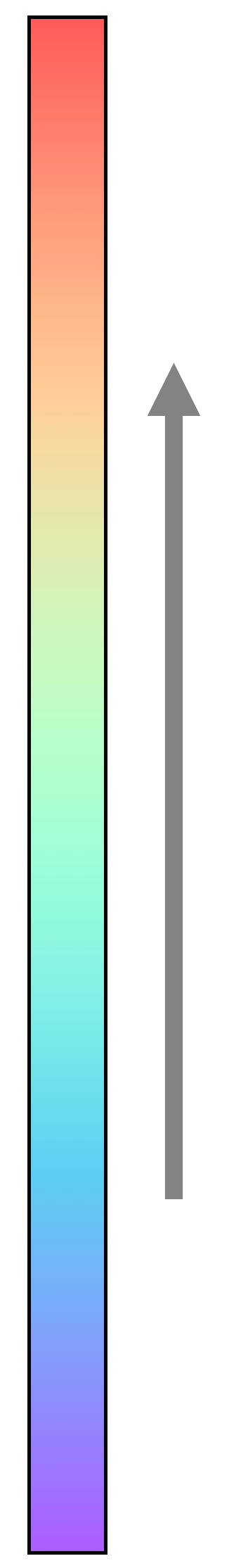}}
            \caption{Anomaly scores $s$ at different time instants $t$, from left to right.}
            \label{fig:anomaly_scores}
            \end{center}
        \end{subfigure}
        \caption{\onlineiforest dynamically adapts to the data distribution of the stream and improves the anomaly scores estimate over time.}
        \label{fig:synthetic}
    \end{figure*}

    \section{Introduction}
        \label{sec:introduction}
        
        Anomaly detection deals with the problem of identifying data that do not conform to an expected behavior~\cite{ChandolaBanerjeeAl2009}. This task finds numerous applications ranging from the financial fraud~\cite{AhmedMahmoodAl2016, DalPozzoloBoracchiAl2018} and intrusion~\cite{BronteShahriarAl2016} detection, to health~\cite{BanaeeAhmedAl2013} and quality monitoring~\cite{StojanovicDinicAl2016}, to name a few examples.
        Usually, anomaly detection methods are trained offline with a static dataset and then used to classify new arriving instances. \textit{Isolation Forest}~\cite{LiuTingAl2008, LiuTingAl2012} (\ifor) is perhaps the most popular offline anomaly detection solution, which is at the same time simple, easy to scale and shows strong performance on a variety of benchmarks.
        
        The popularity of \ifor can be acknowledged by the large number of subsequent works that build on top of it. The majority of these methods aim to improve the performance of \ifor by addressing the limitation of axis-parallel splits~\cite{LiuTingAl2010, HaririKindAl2021, LesoupleBaudoinAl2021, XuPangAl2023}, while others extend \ifor beyond the concept of point-anomaly, to identify functional-anomalies~\cite{StaermanMozharovskyiAl2019} and structured-anomalies~\cite{LeveniMagriAl2021, LeveniMagriAl2023}. Despite their performance, \ifor and its variants are offline anomaly detection methods, and they are unable to operate in a streaming scenario where data comes in endless streams possibly following a dynamic nature. In the online context, offline anomaly detection solutions fail because the models would quickly become outdated and lose performance. The dynamic nature of online environments demands continuous adaptability, making offline approaches unsuitable for effective anomaly detection in the long run.
        
        Online anomaly detection is far less explored compared to online classification~\cite{GomesBarddalAl2017, GomesReadAl2019}, and most of the solutions present in the literature are just online adaptations of offline anomaly detection methods carried out by periodic retraining. Although the retraining approach represents an improvement over the static counterpart, it leads to a substantial processing overhead. Furthermore, online anomaly detection algorithms have to cope with a potentially endless stream of data, strict memory limitations and the single-pass requirement, where it is not possible to store all data for later analysis. Thus, it is essential to have a fast and truly online anomaly detection approach.

        We propose \onlineiforest, an anomaly detection method tailored for the streaming scenario, which can effectively handle the incremental nature of data streams.
        \onlineiforest models the data distribution via an ensemble of multi-resolution histograms endowed with a dynamic mechanism to learn new points and forget old ones. Each histogram collects the points count within its bins and, upon reaching a maximum height, a bin undergoes a split to increase the resolution of the histogram in the most populated regions of the space.
        Conversely, bins in sparsely populated regions are eventually aggregated, thereby decreasing the histogram resolution in the corresponding regions of the space.

        In \cref{fig:synthetic} we illustrate the online learning capabilities of our method with a toy example. Genuine data, depicted in green, are more densely distributed than anomalous one, represented in red. \onlineiforest processes points in~\cref{fig:data} one at a time (i.e., in a streaming fashion), and assigns an anomaly score to each of them. As the stream continues, \onlineiforest acquires more information about the data distribution and refines the estimate of the anomaly scores accordingly (\cref{fig:anomaly_scores}).

        Our experiments demonstrate that \onlineiforest features an exceptionally fast operational speed, addressing the high-speed demands of a streaming context, and achieves effectiveness on par with state-of-the-art online anomaly detection techniques. These properties make \onlineiforest a promising solution for streaming applications. The code of our method is publicly available at \url{https://github.com/ineveLoppiliF/Online-Isolation-Forest}.

    \section{Problem Formulation}
        We address the online anomaly detection problem in a virtually unlimited multivariate data stream $\vect{x}_1, \vect{x}_2, \dots, \vect{x}_t \in \mathbb{R}^d$, where $t \geq 1$. We assume that each $\vect{x}_i$ is a realization of an independently and identically distributed (i.i.d.) random variable having unknown distribution either $\vect{X}_i \sim \Phi_0$ or $\vect{X}_i \sim \Phi_1$, where $\Phi_0$ is the distribution of genuine data and $\Phi_1$ is the distribution of anomalous data.
        Given a point $\vect{x}_t$, the goal is to identify whether $\vect{X}_t \sim \Phi_0$ or $\vect{X}_t \sim \Phi_1$ for each time instant $t$.
        
        We assume that anomalous data are \textit{``few"} and \textit{``different"}~\cite{LiuTingAl2008} and express these assumptions in the following way: (i) \textit{``few"} -- the probability that a point $\vect{x}_i$ has been generated by the distribution $\Phi_1$ of anomalous data is much lower than the probability that it has been generated by the distribution $\Phi_0$ of genuine data, i.e., $P(\vect{X}_i \sim \Phi_1) \ll P(\vect{X}_i \sim \Phi_0)$,
        and (ii) \textit{``different"} -- the probability that a point $\vect{x}_i$ is closer to an anomalous point $\vect{x}_j$ rather than a genuine point $\vect{x}_k$ is low.
        As a consequence, our focus is on scenarios where anomalous data do not form dense and populous clusters.

        Furthermore, we assume that we can store in memory only a finite and small subset $\vect{x}_{t-\omega}, \dots, \vect{x}_t$ of size $\omega$ from the entire data stream at each time instant $t$, were $\omega$ is small enough.
        Additionally, we require the time interval between the acquisition of a sample $\vect{x}_t$ and its classification to be as small as possible.

    \section{Related Work}
        \label{sec:related_works}
        
        Among the wide literature on anomaly detection~\cite{ChandolaBanerjeeAl2009}, we focus on tree-based methods, as they achieve state-of-the-art performance at low computational and memory requirements.
        \ifor~\cite{LiuTingAl2008, LiuTingAl2012} introduced the concept of \textit{``isolation"} as a criteria to categorize anomaly detection algorithms, and subsequent works showed that it is strongly related to the concepts of both distance and density~\cite{ZhangDouAl2017, LeveniMagriAl2023}.
        
        The core component of \ifor is an ensemble of random trees constructed through an iterative branching process. Each individual tree is built by randomly selecting a data dimension and a split value within the bounding box containing data points in that dimension. Anomalous data are identified via an anomaly score computed on the basis of the average path length from the root node to the leaf node, under the assumption that anomalies are easier to isolate.

        The most straightforward extension of \ifor to the streaming scenario is \textit{Isolation Forest ASD}~\cite{DingFei2013} (\asdifor), which periodically trains from scratch a new \ifor ensemble on the most recent data and discards the old ensemble. The periodic retraining delays the adaptation to new data and, depending on how frequently it is performed, slows down the execution. In contrast, \onlineiforest seamlessly updates its internal structure at each new sample processed, enabling fast and truly online anomaly detection.
        \textit{Robust Random Cut Forest}~\cite{GuhaMishraAl2016} (\rrcf) represents the first attempt to adapt \ifor to the streaming context. \rrcf dynamically manages tree structures, and introduces a novel anomaly score based on the discrepancy in tree complexity when a data point is removed from the stream. The anomaly score grounds on the assumption that anomalies' impact on tree structures is more evident compared to genuine data points. However, tree modifications performed by \rrcf tend to be resource-intensive compared to the fast bin splitting and aggregation of \onlineiforest.
        \textit{LODA}~\cite{Pevny2016} leverages on the Johnson–Lindenstrauss~\cite{JohnsonLindenstrauss1984} lemma to project data onto $1$-dimensional spaces and subsequently model data distributions via $1$-dimensional histograms. \loda makes use of fixed resolution histograms, which make it ineffective in describing multi-modal and complex data arrangements. Conversely \onlineiforest histograms, thanks to their ability to increase resolution, successfully adapt to various data configurations.
        \textit{Half Space Trees}~\cite{TanTingAl2011} (\hst), in contrast to \loda, employs an ensemble of $d$-dimensional multi resolution histograms. Specifically, \hst builds an ensemble of complete binary trees by picking a random dimension and using the mid-point to bisect the space. Under the assumption that anomalous data points lie in sparse regions of the space, authors of \hst propose a score based on node masses. Similarly to \asdifor, \hst suffers of periodic retraining. Additionally, the bins in \hst histograms are generated without data, leading to high resolution in empty regions of the space and subsequent memory inefficiency.
        On the other hand, the data-dependent histograms of \onlineiforest allows for a more detailed description of data distribution in the most populous regions of the space.

    \section{Method}
        \label{sec:method}

        \begin{algorithm2e}[t]
            \caption{\onlineiforest}
            \label{alg:online_iforest}
            \DontPrintSemicolon
            \SetNoFillComment
            \KwIn{$\omega$  window size, $\tau$ - number of trees, $\eta$ - max leaf samples}
            initialize $W$ as empty list \label{line:initialize_window}\\
            initialize $\mathcal{F}$ as set of $\tau$ empty trees $\{T_1, \dots, T_{\tau}\}$ \label{line:initialize_forest}\\
            \While{true}
                {$\vect{x}_t \leftarrow$ get point from stream \label{line:get_point}\\
                 \begin{small}
                    \tcc{\textbf{Update forest} \!\!\!}
                 \end{small}
                 append $\vect{x}_t$ to $W$ \label{line:append_window}\\
                 \For{$i = 1$ \normalfont{to} $\tau$ \label{line:insert_loop}}
                     {\texttt{learn\_point}$(\vect{x}_t, T_i.rootN, \eta, \texttt{c}(\omega, \eta))$ \label{line:insert_point}}
                 \If{\normalfont{length of} $W$ \normalfont{greater than} $\omega$ \label{line:check_window}}
                    {$\vect{x}_{t-\omega} \leftarrow$ pop oldest point from $W$ \label{line:remove_window}\\
                     \For{$i = 1$ \normalfont{to} $\tau$ \label{line:remove_loop}}
                         {\texttt{forget\_point}$(\vect{x}_{t-\omega}, T_i.rootN, \eta)$ \label{line:remove_point}}}
                \begin{small}
                    \tcc{\textbf{Score point} \!\!\!}
                \end{small}
                $\mathcal{D} \leftarrow \emptyset$ \label{line:depth_init}\\
                \For{$i = 1$ \normalfont{to} $\tau$ \label{line:score_loop}}
                    {$\mathcal{D} \leftarrow \mathcal{D}$ $\cup$ \texttt{point\_depth}$(\vect{x}_t, T_i.rootN)$  \label{line:compute_depth}}
                $s_t \leftarrow 2^{-\frac{E(\mathcal{D})}{\texttt{c}(\omega, \eta)}}$ \label{line:compute_score}}
        \end{algorithm2e}

        \begin{figure*}[t]
            \begin{subfigure}[b]{0.2\textwidth}
                \begin{center}
                \centerline{\includegraphics[width=\columnwidth]{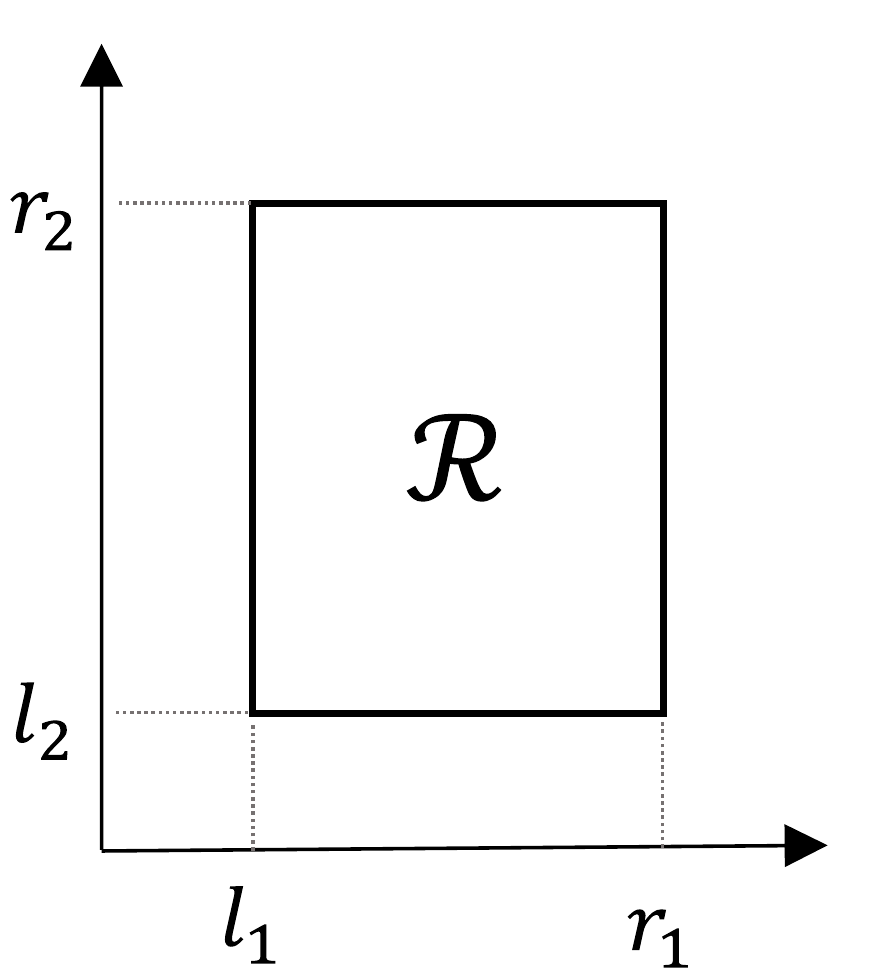}}
                \caption{Bin support $\mathcal{R}$ and its boundaries $[l_i, r_i]$.}
                \label{fig:support}
                \end{center}
            \end{subfigure}
            \hfill
            \begin{subfigure}[b]{0.24\textwidth}
                \begin{center}
                \centerline{\vspace*{0.26cm} \includegraphics[width=\columnwidth]{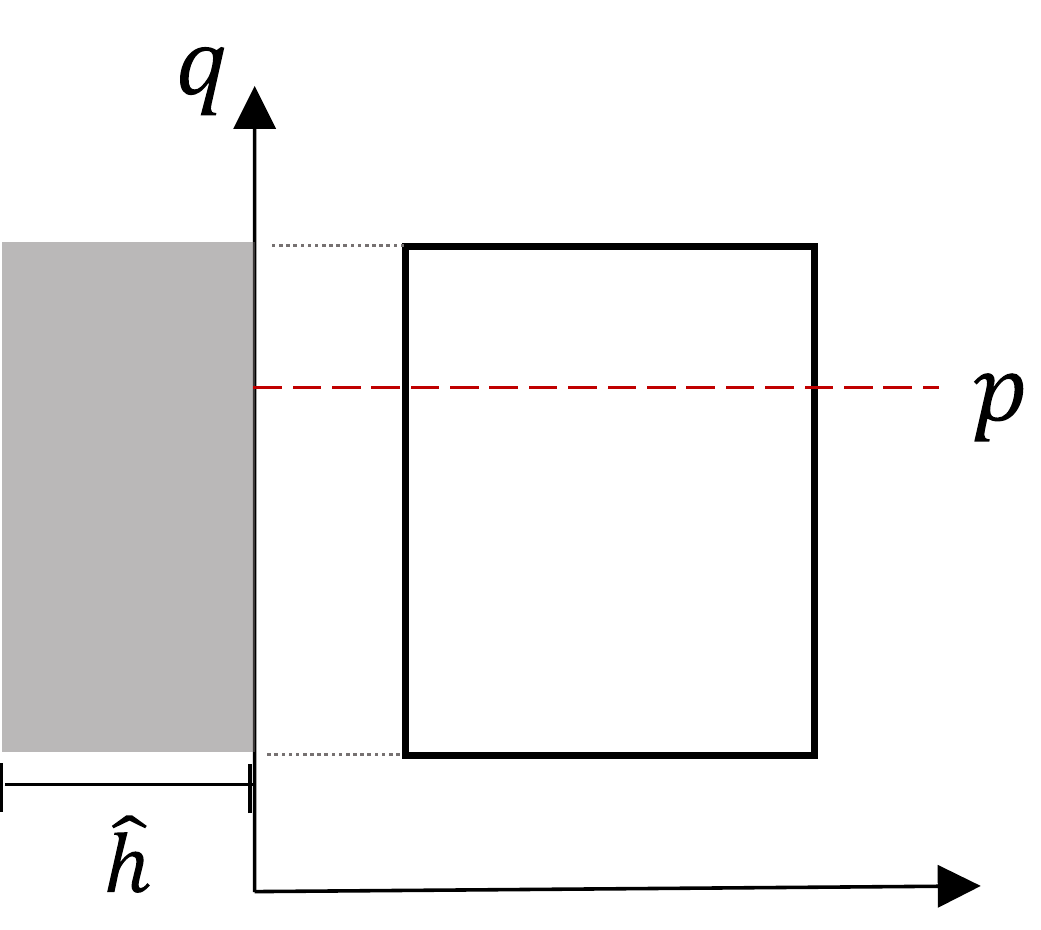}}
                \caption{Maximum bin height $\widehat{h}$ and split information $q$ and $p$.}
                \label{fig:split}
                \end{center}
            \end{subfigure}
            \hfill
            \begin{subfigure}[b]{0.21\textwidth}
                \begin{center}
                \centerline{\vspace*{0.35cm} \includegraphics[width=\columnwidth]{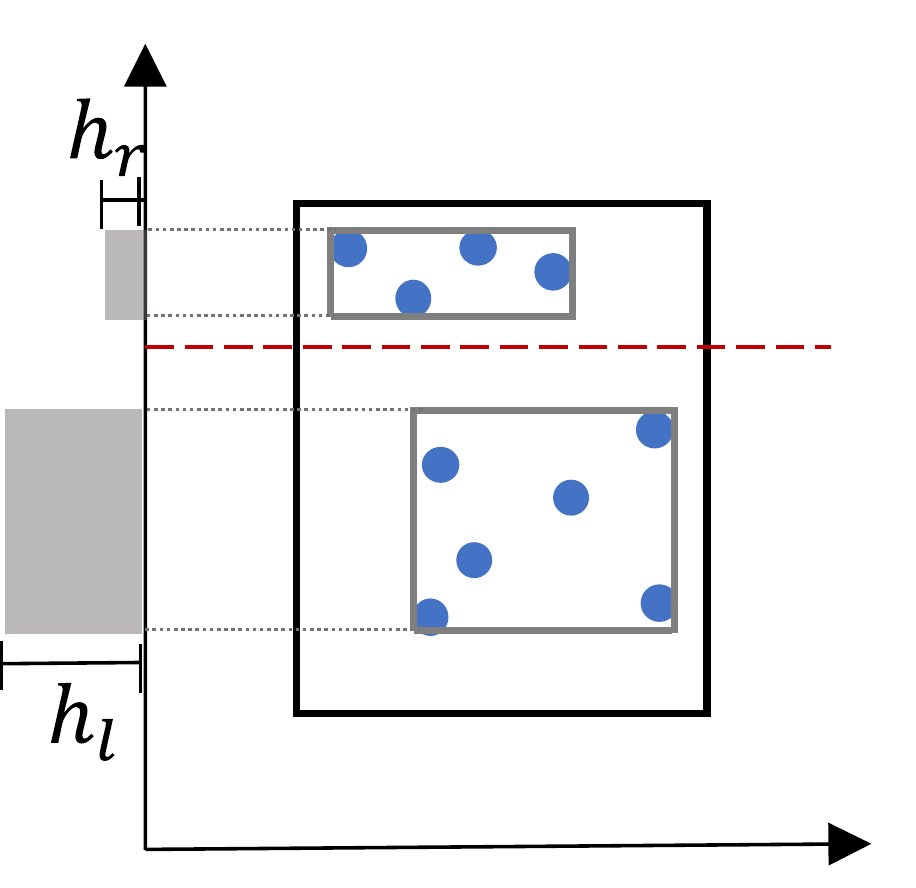}}
                \caption{Sampled points $\mathcal{X}$ and new bins height $h_l$ and $h_r$.}
                \label{fig:points}
                \end{center}
            \end{subfigure}
            \hfill
            \begin{subfigure}[b]{0.185\textwidth}
                \begin{center}
                \centerline{\vspace*{0.35cm} \includegraphics[width=\columnwidth]{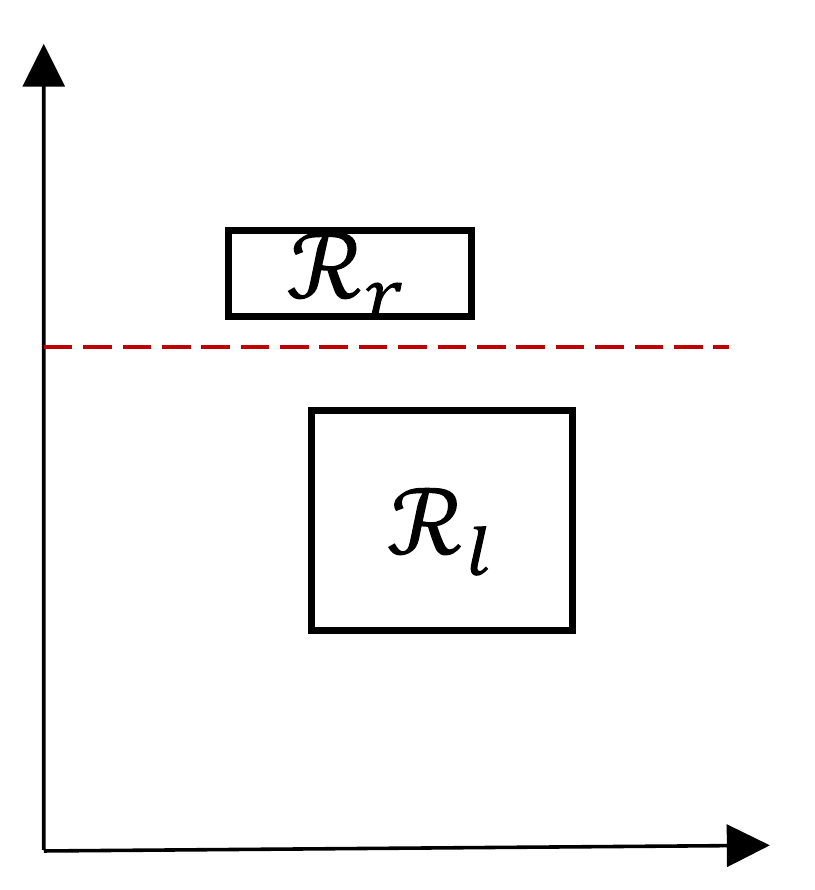}}
                \caption{New bins support $\mathcal{R}_l$ and $\mathcal{R}_r$.}
                \label{fig:new_support}
                \end{center}
            \end{subfigure}
            \caption{Split procedure of a leaf node $N = (h, \mathcal{R})$ in a tree $T$. (b) As soon as a node $N$ reaches the maximum bin height $\widehat{h}$, we randomly select a dimension $q$ and a split value $p$. (c) We randomly sample a set $\mathcal{X}$ of $\widehat{h}$ points from the support $\mathcal{R}$ and use them to initialize bins of newborn child nodes $N_l = (h_l, \mathcal{R}_l)$ and $N_r = (h_r, \mathcal{R}_r)$.}
            \label{fig:split_procedure}
        \end{figure*}
        
        \onlineiforest, denoted as $\mathcal{F}$, is an ensemble of \onlineitrees $\{T_1, \dots, T_\tau\}$ that we specifically designed to continuously and efficiently learn in streaming manner and adapt to the evolving data distribution inherent in streaming contexts.
        Each \onlineitree is a $d$-dimensional histogram that evolves its bins, both in terms of structure and the associated height, as it collects more information about the unknown distributions $\Phi_0$ and $\Phi_1$ over time.
        We rely on a sliding buffer $W = [\vect{x}_{t-\omega}, \dots, \vect{x}_t]$ containing the $\omega$ most recent points and, at each time instant $t$, we use the most and least recent points $\vect{x}_t$ and $\vect{x}_{t-\omega}$ from the buffer $W$ to respectively expand and contract the tree accordingly.
        
        \onlineiforest, detailed in~\cref{alg:online_iforest}, initializes the buffer $W$ as an empty list, and each tree $T \in \mathcal{F}$ is initially composed of the root node only. At each time step $t$, we get a new point $\vect{x}_t$ from the data stream (line~\ref{line:get_point}), store it in the sliding buffer $W$, and use it to update every base model $T$ within our ensemble (lines~\ref{line:append_window}-\ref{line:insert_point}). A tree $T$ learns the new point $\vect{x}_t$ by updating each bin along the path from the root to the corresponding leaf, and potentially expanding its tree structure as described by the learning procedure in~\cref{alg:online_itree_insert}. Subsequently, we remove the oldest point $\vect{x}_{t-\omega}$ from the buffer $W$ and force every tree $T$ to forget it (lines~\ref{line:check_window}-\ref{line:remove_point}) via the forgetting procedure that entails updating the bins along the leaf path. The forgetting step, in contrast to the learning one, involves a potential contraction of the tree structure instead of an expansion, as outlined in~\cref{alg:online_itree_remove}. We formalize and detail the learning and forgetting procedures in ~\cref{subsec:online_itree}.
        
        We compute the anomaly score $s_t \in [0, 1]$ for the point $\vect{x}_t$ (lines~\ref{line:depth_init}-\ref{line:compute_score}) according to the principles behind \ifor~\cite{LiuTingAl2008}. In particular, anomalous points are easier to isolate than genuine ones, and therefore they are more likely to be separated early in the recursive splitting process of a random tree.
        Therefore, we determine the anomaly score by computing the depth of all the leaves where $\vect{x}_t$ falls in each tree $T \in \mathcal{F}$, and consequently mapping the depths to the corresponding $s_t$ via the following normalization function
        \begin{equation*}
            s_t \leftarrow 2^{-\frac{E(\mathcal{D})}{\texttt{c}(\omega, \eta)}},
        \end{equation*}
        where $E(\mathcal{D})$ is the average depth along all the $\tau$ trees of the ensemble, and $\texttt{c}(\omega, \eta) = \log_2\frac{\omega}{\eta}$ is an adjustment factor as a function of the window size $\omega$ and the number $\eta$ of points required to perform a bin split. The adjustment factor represents the average depth of an \onlineitree and we derive it in~\cref{subsec:complexity_analysis}.
        The computation of the leaf depth where $\vect{x}_t$ falls into is detailed in~\cref{alg:online_itree_depth}.
        
        \subsection{\onlineitree}
            \label{subsec:online_itree}

            The fundamental component of our solution is the \onlineitree structure. Every \onlineitree is a $d$-dimensional histogram constructed by recursively splitting the input space $\mathbb{R}^d$ into bins, such that each bin stores the number of points that fell in the corresponding region of the space. We define \onlineitree as a dynamic collection of nodes $T = \{N_j\}_{j = 1, \dots, m}$ that is continuously updated as new points are learned and old points are forgotten by the tree. We characterize the $j$-th node as $N_j = (h_j, \mathcal{R}_j)$, where $h_j$ is the number of points that crossed it in their path to the leaf, that is the bin height, and $\mathcal{R}_j = \bigtimes_{i = 1}^d [l_i, r_i]$ is the minimal $d$-dimensional hyperrectangle that encloses them, that is the support of the bin, where $\bigtimes$ denotes the Cartesian product.
            It is worth noting that $h_j$ and $\mathcal{R}_j$ are sufficient for achieving an efficient online adaptation of \ifor.

            When a new sample $\vect{x}_t$ is received from the data stream we run, independently on each \onlineitree, a learning procedure to update the tree. The learning procedure involves sending the incoming sample $\vect{x}_t$ to the corresponding leaf, and updating the heights $h$ and supports $\mathcal{R}$ of all the bins along the path accordingly.
            When a leaf reaches the maximum height $\widehat{h}$, we split the corresponding bin in two according to the procedure illustrated in~\cref{fig:split_procedure} and described next.
            This is repeated until the window $W$ gets full, then, together with the learning procedure for the new incoming sample $\vect{x}_t$, we include a forgetting procedure for the oldest sample $\vect{x}_{t-\omega}$ in $W$. The forgetting procedure might involve aggregating nearby bins in a single one as illustrated in~\cref{fig:forget_procedure}. The two fold updating mechanism enables \onlineitree to (i) incrementally learn when the stream starts and (ii) track possible evolution of the stream.
            
            \onlineitree bins associated to more populated regions of the space undergo frequent splits, whereas bins associated with sparsely populated regions undergo less frequent splits. Since each split increases the depth of the tree's leaf nodes, we can distinguish between anomalous and genuine points based on the depth $k$ of the leaf nodes they fall into. In~\cref{fig:anomaly_scores} we see that an ensemble of \onlineitrees assigns different anomaly scores to regions with high and low population.

            \subsubsection*{Learning procedure}
                \label{subsubsec:learn_procedure}

                \begin{algorithm2e}[t]
                    \caption{\onlineitree{} -- learn point}
                    \label{alg:online_itree_insert}
                    \DontPrintSemicolon
                    \SetNoFillComment
                    \KwIn{$\vect{x}$ - input data, $N$ - a tree node, $\eta$ - max leaf samples, $\delta$ - depth limit}
                    \SetKwFunction{FMain}{learn\_point}
                    \SetKwProg{Fn}{Function}{:}{}
                    \Fn{\FMain{$\vect{x}, N, \eta, \delta$}}{
                    update bin height $h$ and support $\mathcal{R}$ \label{line:update_bin} \\
                    \If{$N$ \normalfont{is a leaf}}
                        {\If{$h \geq \eta \: 2^{k}$ \normalfont{\textbf{and}} $k < \delta$ \label{line:split_condition}}
                            {$q \leftarrow$ sample from $\mathcal{U}_{\{1, \dots, d\}}$ \label{line:sample_dimension} \\
                             $p \leftarrow$ sample from $\mathcal{U}_{[l_q, r_q]}$ \label{line:sample_value} \\
                             $\mathcal{X} \leftarrow$ sample from $\mathcal{U}_{\mathcal{R}}$ \label{line:sample_points} \\
                             partition $\mathcal{X}$ into $\mathcal{X}_l$, $\mathcal{X}_r$ \label{line:split_points} \\
                             compute $h_l$, $h_r$ and $\mathcal{R}_l$, $\mathcal{R}_r$ \\
                             initialize child nodes $N_l, N_r$ \label{line:initialize_children} 
                         }
                       \Else
                        {\If{$x_q < p$}
                            {\texttt{learn\_point}$(\vect{x}, N_l, \eta, \delta)$}
                         \Else
                            {\texttt{learn\_point}$(\vect{x}, N_r, \eta, \delta)$}
                        }}}
                \end{algorithm2e}

                \begin{figure}[t]
                    \begin{subfigure}[b]{0.235\textwidth}
                        \begin{center}
                        \centerline{\includegraphics[width=\columnwidth]{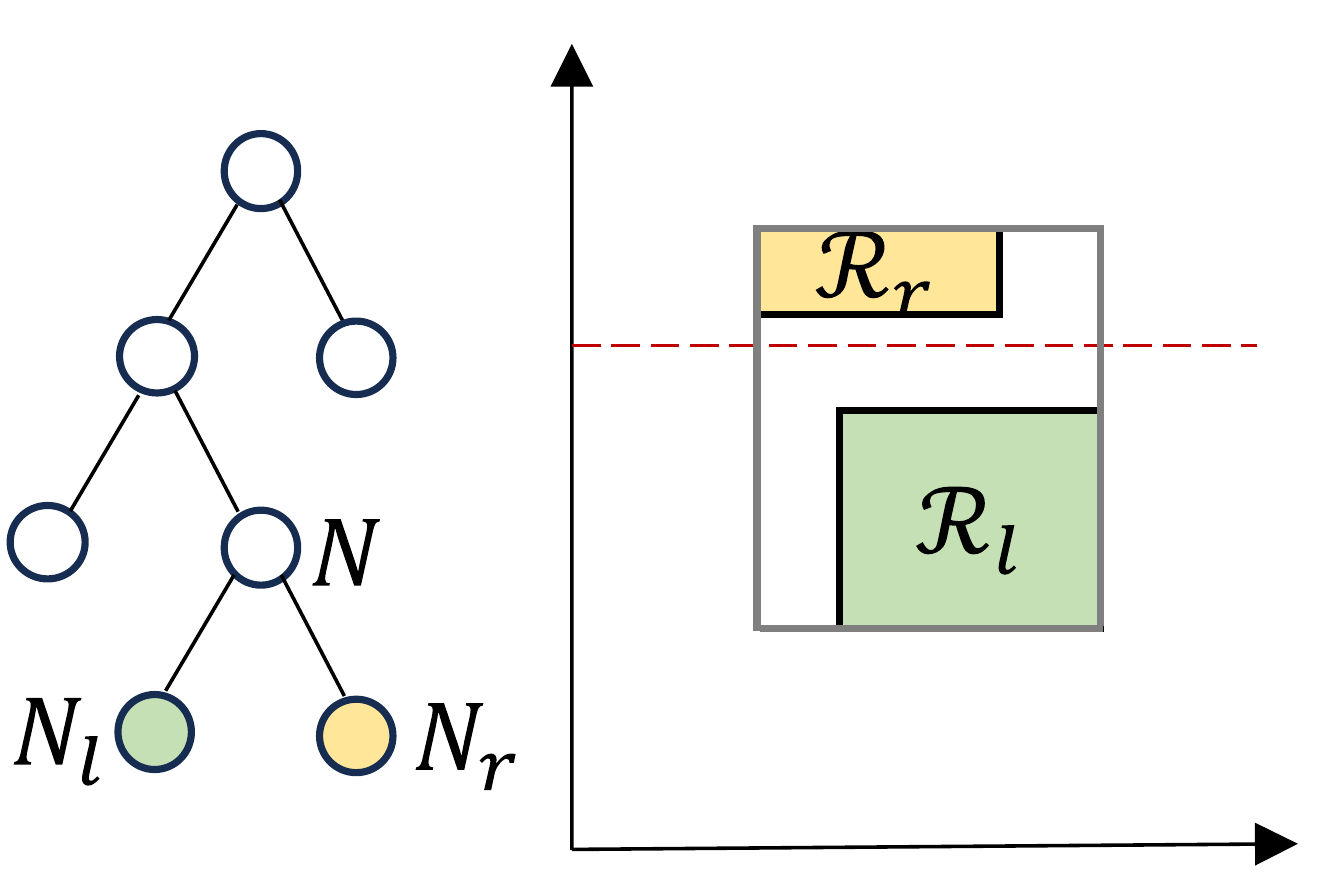}}
                        \caption{Before}
                        \label{fig:before_forget}
                        \end{center}
                    \end{subfigure}
                    \hfill
                    \begin{subfigure}[b]{0.235\textwidth}
                        \begin{center}
                        \centerline{\includegraphics[width=\columnwidth]{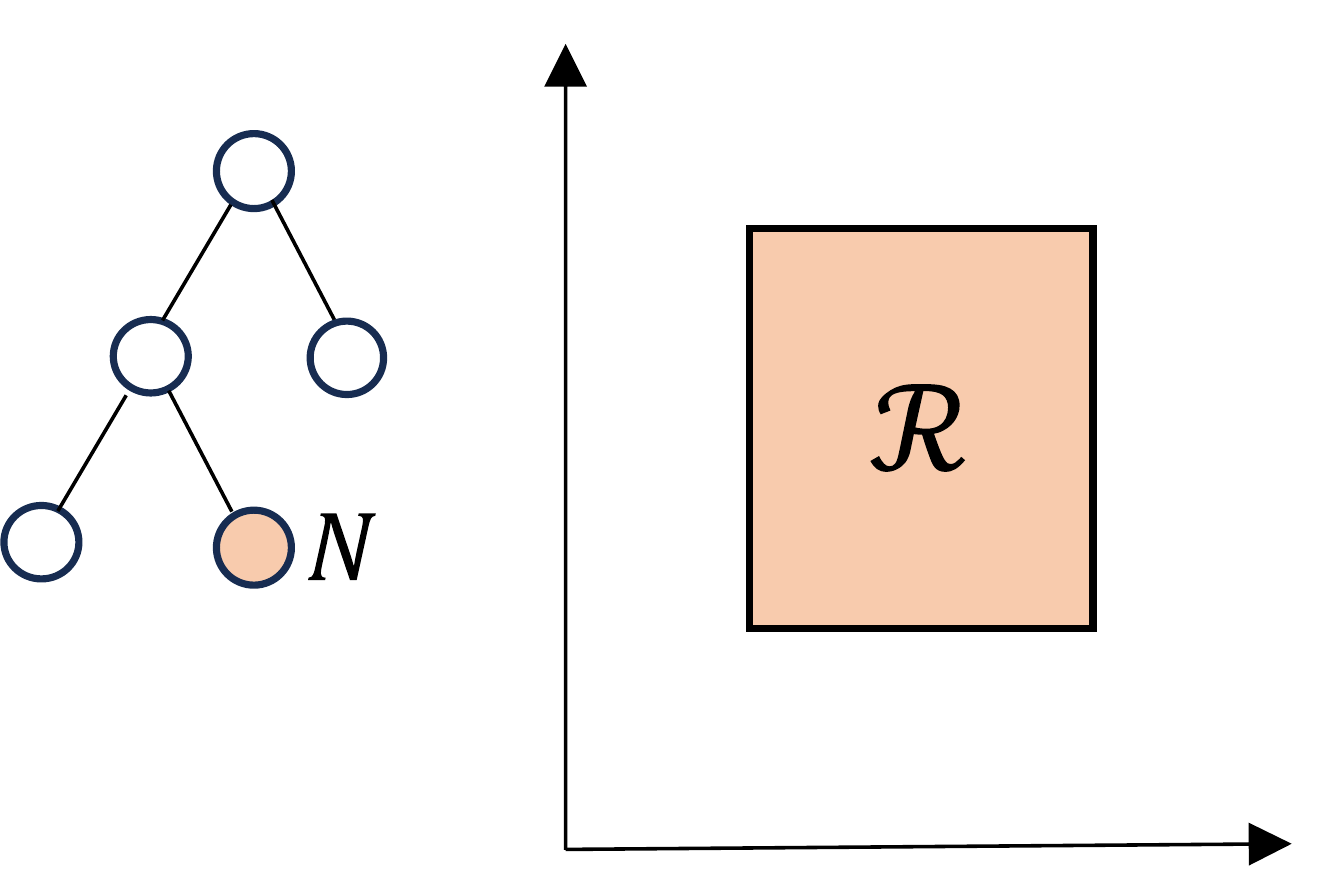}}
                        \caption{After}
                        \label{fig:after_forget}
                        \end{center}
                    \end{subfigure}
                    \caption{Following the forgetting procedure, support $\mathcal{R}$ of node $N$ is the minimal hyperrectangle that encloses supports $\mathcal{R}_l, \mathcal{R}_r$ of child nodes $N_l, N_r$.}
                    \label{fig:forget_procedure}
                \end{figure}

                \begin{algorithm2e}[t]
                     \caption{\onlineitree{} -- forget point}
                     \label{alg:online_itree_remove}
                     \DontPrintSemicolon
                     \SetNoFillComment
                     \KwIn{$\vect{x}$ - input data, $N$ - a tree node, $\eta$ - max leaf samples}
                     \SetKwFunction{FMain}{forget\_point}
                     \SetKwProg{Fn}{Function}{:}{}
                     \Fn{\FMain{$\vect{x}, N, \eta$}}{
                     decrease bin height $h$ \label{line:update_bin_remove} \\
                     \If{$N$ \normalfont{is NOT a leaf}}
                             {\If{$h < \eta \: 2^{k}$ \label{line:forget_condition}}
                                 {update bin support $\mathcal{R}$ from $\mathcal{R}_l, \mathcal{R}_r$\label{line:update_support} \\
                                  forget split $q, p$ and child nodes $N_l, N_r$ \label{line:forget_split}
                                  }
                                  
                             \Else
                                 {\If{$x_q < p$}
                                     {\texttt{forget\_point}$(\vect{x}, N_l, \eta)$}
                                  \Else
                                     {\texttt{forget\_point}$(\vect{x}, N_r, \eta)$}
                                 }
                             }
                     }
                \end{algorithm2e}

                Every time we feed a point $\vect{x}_t$ to the tree via the learning procedure, we update both the height $h$ and the support $\mathcal{R}$ of all the nodes crossed by $\vect{x}_t$ along the path from the root to the leaf (line~\ref{line:update_bin} of \cref{alg:online_itree_insert}). When the bin height $h$ of a leaf node $N$ reaches a maximum value $\widehat{h}$, we increase the resolution of the histogram in the corresponding region of the space by splitting the associated bin in two, and generating two child nodes $N_l$ and $N_r$.
                
                We define the maximum height of a bin as $\widehat{h} = \eta \: 2^k$, where $\eta$ is a user-defined parameter, and $k$ is the depth of the corresponding node $N \in T$. Since the maximum height $\widehat{h}$ grows exponentially as a function of the depth $k$ of the considered node, we increase the number of points required to perform a split in deeper nodes of the tree. This design choice results in compact trees, leading to a substantial efficiency gain.
                We perform the split procedure only if the depth $k$ of the leaf node $N$ is less than a maximum value $\delta = \log_2\frac{\omega}{\eta}$, as a function of the window size $\omega$ and the number $\eta$ of points required to split histogram bins. Trees with limited depth had already been discussed in \ifor as a solution to the swamping and masking effects~\cite{Murphy1951} in anomaly detection.
    
                The split procedure, invoked when the bin height $h$ reaches the maximum value $\widehat{h}$, is as follows. First, we randomly sample a split dimension $q \in \{1, \dots, d\}$ and split value $p \in [l_q, r_q]$ from the random variables $Q \sim \mathcal{U}_{\{1, \dots, d\}}$ and $P \sim \mathcal{U}_{[l_q, r_q]}$ respectively, where $\mathcal{U}$ denotes the uniform distribution (lines~\ref{line:sample_dimension}-\ref{line:sample_value} of \cref{alg:online_itree_insert}).
                Second, we sample a set $\mathcal{X}$ of $\widehat{h}$ points from the support $\mathcal{R} = \bigtimes_{i = 1}^d [l_i, r_i]$ such that each element $\vect{x} \in \mathcal{X}$ is distributed according to $\vect{X} \sim \mathcal{U}_{\mathcal{R}}$ (line~\ref{line:sample_points}). The uniform sampling of $\mathcal{X}$ grounds on the approximation of the data distribution by a piece wise uniform distribution represented by the union of histogram leaves' bins.
                Finally, we partition the elements of $\mathcal{X}$ into $\mathcal{X}_l = \{\vect{x} \in \mathcal{X} | x_q < p\}$ and $\mathcal{X}_r = \{\vect{x} \in \mathcal{X} | x_q \geq p\}$, and use them to initialize heights $h_l$, $h_r$ and supports $\mathcal{R}_l$, $\mathcal{R}_r$ of the newborn left and right child nodes $N_l$ and $N_r$ respectively (lines~\ref{line:split_points}-\ref{line:initialize_children}).
                For illustration purposes, we depicted the split procedure in~\cref{fig:split_procedure} when $d = 2$.

            \subsubsection*{Forgetting procedure}
                \label{subsubsec:forget_procedure}

                \begin{algorithm2e}[t]
                    \caption{\onlineitree{} -- point depth}
                    \label{alg:online_itree_depth}
                    \DontPrintSemicolon
                    \SetNoFillComment
                    \KwIn{$\vect{x}$ - input data, $N$ - a tree node, $\eta$ - max leaf samples}
                    \KwOut{depth of point $\vect{x}$}
                    \SetKwFunction{FMain}{point\_depth}
                    \SetKwProg{Fn}{Function}{:}{}
                    \Fn{\FMain{$\vect{x}, N$}}{
                    \If{$N$ \normalfont{is a leaf}}
                       {\Return $k + \texttt{c}(h, \eta)$}
                    \Else
                        {\If{$x_q < p$}
                            {\Return \texttt{point\_depth}$(\vect{x}, N_l)$}
                         \Else
                            {\Return \texttt{point\_depth}$(\vect{x}, N_r)$}}
                    }
                \end{algorithm2e}

                In contrast to the learning procedure, which involves creating new nodes and thereby enhancing the histogram resolution in that area, in the forgetting procedure we aggregate nodes and merge the associated bins, ultimately reducing the number of bins and, hence, the histogram resolution in the corresponding region of the space. Specifically, every time we feed a point $\vect{x}_{t-\omega}$ to the tree via the forgetting procedure, we decrease the bin height $h$ of all the nodes $N$ crossed by it along the path from the root to the leaf (line~\ref{line:update_bin_remove} of \cref{alg:online_itree_remove}). When the height $h$ of an internal node (i.e., of a node that experienced a split) drops below the threshold $\widehat{h}$, we forget the split in $N$ by merging its two child nodes $N_l$ and $N_r$.
                The forget procedure (illustrated in~\cref{fig:forget_procedure}) consists in first updating the bin support $\mathcal{R}$ of the node $N$ as the minimal hyperrectangle that encloses bin supports $\mathcal{R}_l$ and $\mathcal{R}_r$ (line~\ref{line:update_support} of \cref{alg:online_itree_remove}) of $N_l$ and $N_r$ respectively. Then split information $q, p$ and child nodes $N_l, N_r$ are discarded (line~\ref{line:forget_split}).
            
        \subsection{Complexity Analysis}
            \label{subsec:complexity_analysis}

            The computational complexity is a crucial aspect in the online context, where data streams must be processed at high speed with low memory requirements. Time and space complexities of \onlineiforest are closely tied to the depth of the \onlineitrees within the ensemble. Therefore, we first derive \onlineitree depth in both the average and worst case scenarios, and then express time and space complexities of \onlineiforest as functions of these depths (\cref{tab:complexity}).
            
            \paragraph{Average case} To determine the average depth $\bar{k}$ of an \onlineitree, we first note that a perfectly balanced binary tree constructed with $\omega$ points has exactly $\frac{\omega}{2^{k}}$ points at each node at depth $k$~\cite{Knuth2023}. Since a node at depth $k$ requires $\eta \: 2^{k}$ points to undergo a split in \onlineitree, we can state that depth $k$ exists if and only if
            \begin{equation*}
                \frac{\omega}{2^{k-1}} \geq \eta \: 2^{k-1},
            \end{equation*}
            i.e., if there were enough points at depth $k - 1$ to perform the split.
            Making explicit the inequality with respect to $k$ we have
            \begin{equation*}
                k \leq \frac{1}{2} \: \log_2\frac{\omega}{\eta} + 1,
            \end{equation*}
            from which it follows that the average depth of an \onlineitree is
            \begin{equation*}
                \bar{k} = \lfloor \frac{1}{2} \: \log_2\frac{\omega}{\eta} + 1 \rfloor.
            \end{equation*}
            Hence, the average \textit{time} complexity of \onlineiforest (i.e., the computational complexity of traversing each tree from the root to a leaf), is $O(n \: \tau \: \log_2\frac{\omega}{\eta})$, where $\tau$ it the number \onlineitrees in the ensemble and $n$ is the number of samples in the data stream.
            We express the average \textit{space} complexity of \onlineiforest (i.e., the amount of memory space required) as a function of the number of nodes in an \onlineitree, that is in turn tied to the average depth $\bar{k}$, plus the buffer size $\omega$. Specifically, we note that the number of nodes in a perfectly balanced binary tree with depth $k$ is $2^{k+1} - 1$~\cite{Knuth2023}. Therefore, the average space complexity of \onlineiforest is $O(\tau \: 2^{\frac{1}{2}\log_2\frac{\omega}{\eta}} + \omega) = O(\tau \sqrt{\frac{\omega}{\eta}} + \omega)$, and it is independent from the number $n$ of samples in the data stream.

            \begin{table}[t]
                \caption{Average and worst case complexity of \onlineiforest.}
                \begin{center}
                    \begin{small}
                    \begin{sc}
                    \begin{tabular}{lcc}
                    \toprule
                    Complexity & Average case                                  & Worst case                             \\
                    \midrule
                    Time       & $O(n \: \tau \: \log_2\frac{\omega}{\eta})$   & $O(n \: \tau \: \log_2\frac{w}{\eta})$ \\
                    Space      & $O(\tau \sqrt{\frac{\omega}{\eta}} + \omega)$ & $O(\tau \frac{\omega}{\eta} + \omega)$ \\
                    \bottomrule
                    \end{tabular}
                    \end{sc}
                    \end{small}
                \end{center}
                \label{tab:complexity}
            \end{table}

            \paragraph{Worst case} We note that a binary tree degenerated into a linked list, constructed with $\omega$ points, has $\omega - k$ points at depth $k$. Therefore, similarly to the average case, depth $k$ exists if and only if
            \begin{equation*}
                \omega-(k-1) \geq \eta \: 2^{k-1}.
            \end{equation*}
            By making the depth $k$ explicit, and placing an upper bound on it, we have
            \begin{equation*}
                k \leq \log_2\frac{w+1-k}{\eta} + 1 \leq \log_2\frac{w+1}{\eta} + 1
            \end{equation*}
            from which it follows that the worst case depth of an \onlineitree is
            \begin{equation*}
                 \tilde{k} \leq \lfloor \log_2\frac{w+1}{\eta} + 1 \rfloor.
            \end{equation*}
            Thus, the worst case \textit{time} and \textit{space} complexities of \onlineiforest are $O(n \: \tau \: \log_2\frac{w}{\eta})$ and $O(\tau \: 2^{\log_2\frac{w+1}{\eta}} + \omega) = O(\tau \frac{\omega}{\eta} + \omega)$ respectively.

        \subsection{Adaptation Speed vs. Modeling Accuracy}
            \label{subsec:sliding_buffer}

            The length $\omega$ of the sliding buffer $W$ plays a crucial role in controlling the trade-off between adaptation speed and modeling accuracy in \onlineiforest. Specifically, adopting a small $\omega$ allows \onlineiforest to quickly adapt to changes, but it results in a coarse modeling of the underlying data distribution. To this regard, we can observe~\cref{fig:anomaly_scores}, illustrating the anomaly scores as \onlineiforest processes an increasing number of points. Moving from left to right, the anomaly scores describe the learned data distribution after processing $100$, $300$ and $1000$ points, and this is equivalent to what \onlineiforest would learn over sliding buffers of corresponding lengths. Notably, after processing $100$ points, the anomaly scores are coarse, whereas they become more fine-grained after $1000$ points.

    \section{Experiments}
        \label{sec:experiments}
    
        In this section we first compare the performance of \onlineiforest and state-of-the-art methods on a large anomaly detection benchmark where anomalous and genuine distributions $\Phi_0$ and $\Phi_1$ are stationary. Then, we resort to a dataset exhibiting concept drift to assess their capability to adapt to distribution changes.
    
        \subsection{Datasets}
            \label{subsec:datasets}

            \begin{table}[t]
                \vskip 0.15in
                \caption{Stationary datasets properties.}
                \begin{center}
                    \begin{small}
                    \begin{sc}
                    \begin{tabular}{lrrrr}
                    \toprule
                    Dataset              & $n$      & $d$  & $\%$ of anomalies \\
                    \midrule
                    Donors               & $619326$ & $10$ & $5.90$            \\
                    Http                 & $567497$ & $3$  & $0.40$            \\
                    ForestCover          & $286048$ & $10$ & $0.90$            \\
                    fraud                & $284807$ & $29$ & $0.17$            \\
                    Mulcross             & $262144$ & $4$  & $10.00$           \\
                    Smtp                 & $95156$  & $3$  & $0.03$            \\
                    Shuttle              & $49097$  & $9$  & $7.00$            \\
                    Mammography          & $11183$  & $6$  & $2.00$            \\
                    nyc\_taxi\_shingle   & $10273$  & $48$ & $5.20$            \\
                    Annthyroid           & $6832$   & $6$  & $7.00$            \\
                    Satellite            & $6435$   & $36$ & $32.00$           \\
                    \bottomrule
                    \end{tabular}
                    \end{sc}
                    \end{small}
                \end{center}
                \vskip -0.1in
                \label{tab:datasets}
            \end{table}

            \paragraph{Stationary} We run our experiments on the eight largest datasets used in~\cite{LiuTingAl2008, LiuTingAl2012} (Http, Smtp~\cite{YamanishiTakeuchiAl2004}, Annthyroid, Forest Cover Type, Satellite, Shuttle~\cite{AsunctionNewman2007}, Mammography and Mulcross~\cite{RockeWoodruff1996}), two datasets from Kaggle competitions (Donors and Fraud~\cite{PangShenAl2019}), and the shingled version of NYC Taxicab dataset used in~\cite{GuhaMishraAl2016}.
            We chose these datasets as they contain real data where the genuine and anomalous distributions $\Phi_0$ and $\Phi_1$ are unknown, and contain labels about anomalous data to perform performance evaluation.
            Information on the cardinality $n$, dimensionality $d$, and $\%$ of anomalies for the datasets is outlined in~\cref{tab:datasets}.

            \paragraph{Non-stationary} We use the INSECTS dataset~\cite{SouzaDosReisAl2020} previously used for change detection~\cite{FrittoliCarreraAl2022, StucchiMagriAl2023, StucchiRizzoAl2023}. INSECTS contains feature vectors ($d = 33$) describing the wing-beat frequency of six (annotated) species of flying insects. This dataset contains $5$ real changes caused by temperature modifications that affect the insects’ flying behavior. For our purposes, we selected the two most populous classes as genuine (`ae-aegypti-female', `cx-quinq-male'), and the least populous as the anomalous one (`ae-albopictus-male'). This results in a total of $n = 212514$ points with $5.50\%$ of anomalies.

        \subsection{Competing methods and methodology}
            \label{subsec:competing_methods}

            In our experiments we compared \onlineiforest (\oifor) to state-of-the-art methods in the online anomaly detection literature described in~\cref{sec:related_works}. In particular, we compared to \textit{iForestASD} (\asdifor), \textit{Half Space Trees} (\hst), \textit{Robust Random Cut Forest} (\rrcf) and \textit{LODA} (\loda) using their \textit{PySAD}~\cite{YilmazKozat2020} implementation.
            
            For comparison purposes, we set the number of trees $\tau = 32$ for all the algorithms, and considered the number of random cuts in \loda equivalent to the number of trees. We set window size $\omega = 2048$ for both \oifor and \asdifor, and used the default value $\omega = 250$ for \hst. We set the subsampling size used to build trees in \asdifor to the default value $\psi = 256$, while the number of bins for each random projection in \loda to $b = 100$. The trees maximum depth $\delta$ depends on the subsamping size $\psi$ in \asdifor, on the window size $\omega$ and number $\eta$ of points required to split histogram bins in \oifor, while it is fixed to the default value $\delta = 15$ in \hst. The parameters configuration for all the algorithms is illustrated in~\cref{tab:parameters}.

            Each algorithm was executed $30$ times on both stationary and non-stationary datasets. We randomly shuffled every stationary dataset before each execution, then used the same shuffled version to test all the algorithms.
            Processing each data point individually is prohibitive in terms of time due to the data stream size. To solve this problem we divided every dataset in batches of $100$ points each and passed one chunk at a time to the algorithms in an online manner.

            We use the ROC AUC and execution time (in seconds) to evaluate the effectiveness and efficiency of the considered algorithms, respectively. In addition, we employ the critical difference diagram for both metrics to synthesize the results across multiple executions on datasets with diverse characteristics.

            \begin{table}[t]
                \vskip 0.15in
                \caption{Algorithms execution parameters.}
                \begin{center}
                    \begin{small}
                    \begin{sc}
                    \begin{tabular}{lcccccc}
                    \toprule
                    Algorithm     & $\tau$ & $\omega$ & $\eta$ & $\psi$ & $\delta$       & $b$   \\
                    \midrule
                    \oifor        & $32$   & $2048$   & $32$   & --     & $\log_2\frac{\omega}{\eta}$ & --    \\
                    \asdifor      & $32$   & $2048$   & --     & $256$  & $\log_2\psi$   & --    \\
                    \hst          & $32$   & $250$    & --     & --     & $15$         & --    \\
                    \rrcf         & $32$   & --       & --     & $256$  & --             & --    \\
                    \loda         & $32$   & --       & --     & --     & --             & $100$ \\
                    \bottomrule
                    \end{tabular}
                    \end{sc}
                    \end{small}
                \end{center}
                \vskip -0.1in
                \label{tab:parameters}
            \end{table}
            
        \subsection{Results}
            \label{subsec:results}

            \begin{table*}[t]
                \hspace*{0.5cm}
                \caption{ROC AUCs and total execution times.}
                \begin{subtable}{\columnwidth}
                    \begin{center}
                        \begin{small}
                        \begin{sc}
                        \tabcolsep=0.15cm
                        \begin{tabular}{lrrrrrrrrrrrr}
                                                     && \multicolumn{5}{c}{AUC $(\uparrow)$}                                         && \multicolumn{5}{c}{Time $(\downarrow)$}                          \\
                                                        \cmidrule{3-7}                                                                  \cmidrule{9-13}
                                                     && oIFOR            & asdIFOR          & HST       & RRCF             & LODA    && oIFOR             & asdIFOR  & HST       & RRCF      & LODA      \\
                            \midrule
                            Donors                   && $\mathbf{0.795}$ & $0.769$          & $0.715$   & $0.637$          & $0.554$ && $\mathbf{252.36}$ & $551.85$ & $2145.85$ & $4924.46$ & $2111.09$ \\
                            Http                     && $0.998$          & $\mathbf{0.999}$ & $0.992$   & $0.996$          & $0.632$ && $\mathbf{179.36}$ & $509.40$ & $2016.00$ & $8367.16$ & $2017.85$ \\
                            ForestCover              && $0.887$          & $0.861$          & $0.722$   & $\mathbf{0.917}$ & $0.500$ && $\mathbf{107.65}$ & $197.82$ & $1045.39$ & $2997.86$ & $1009.92$ \\
                            fraud                    && $0.936$          & $0.946$          & $0.910$   & $\mathbf{0.951}$ & $0.722$ && $\mathbf{100.09}$ & $285.69$ & $973.91$  & $4936.03$ & $931.93$  \\
                            Mulcross                 && $\mathbf{0.995}$ & $0.952$          & $0.011$   & $0.800$          & $0.506$ && $\mathbf{90.33}$ & $270.79$ & $936.01$  & $3244.96$ & $848.12$  \\
                            Smtp                     && $0.861$          & $\mathbf{0.905}$ & $0.851$   & $0.894$          & $0.731$ && $\mathbf{29.95}$  & $142.65$ & $325.77$  & $1273.98$ & $254.69$  \\
                            Shuttle                  && $0.992$          & $\mathbf{0.996}$ & $0.981$   & $0.957$          & $0.528$ && $\mathbf{16.35}$  & $108.28$ & $167.48$  & $770.61$  & $130.61$  \\
                            Mammography              && $0.854$          & $\mathbf{0.855}$ & $0.831$   & $0.824$          & $0.622$ && $\mathbf{3.32}$   & $80.01$  & $37.92$   & $118.22$  & $29.55$   \\
                            nyc\_taxi\_shingle       && $0.572$          & $0.709$          & $0.342$   & $\mathbf{0.725}$ & $0.499$ && $\mathbf{8.03}$   & $83.15$  & $36.70$   & $151.67$  & $36.82$   \\
                            Annthyroid               && $0.685$          & $\mathbf{0.810}$ & $0.636$   & $0.740$          & $0.589$ && $\mathbf{2.00}$   & $77.06$  & $24.26$   & $93.40$   & $19.79$   \\
                            Satellite                && $0.651$          & $\mathbf{0.709}$ & $0.531$   & $0.662$          & $0.501$ && $\mathbf{3.74}$   & $78.90$  & $21.78$   & $93.77$   & $17.55$   \\
                            \midrule
                            median                   && $\mathbf{0.866}$ & $0.863$          & $0.739$   & $0.832$          & $0.541$ && $\mathbf{29.95}$  & $142.57$ & $323.29$  & $1274.45$ & $254.64$  \\
                            mean rank                && $2.167$          & $\mathbf{1.583}$ & $3.917$   & $2.500$          & $4.833$ && $\mathbf{1.000}$   & $2.667$   & $3.500$    & $5.000$    & $2.833$    \\
                            \bottomrule
                        \end{tabular}
                        \end{sc}
                        \end{small}
                    \end{center}
                    \end{subtable}
                \label{tab:performance}
            \end{table*}

            \begin{figure*}[t]
                \begin{subfigure}[b]{\columnwidth}
                    \begin{center}
                    \centerline{\includegraphics[width=.9\columnwidth]{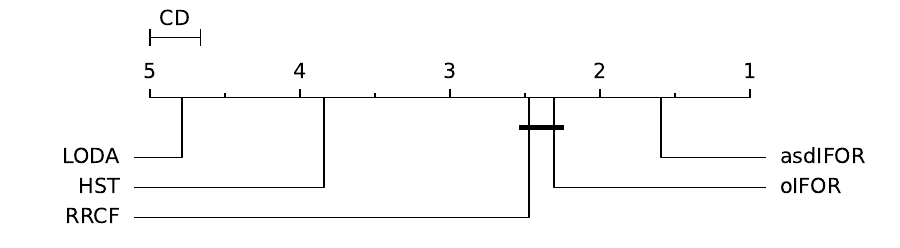}}
                    \caption{ROC AUCs}
                    \label{fig:cd_roc_aucs}
                    \end{center}
                \end{subfigure}
                \begin{subfigure}[b]{\columnwidth}
                    \begin{center}
                    \centerline{\includegraphics[width=.9\columnwidth]{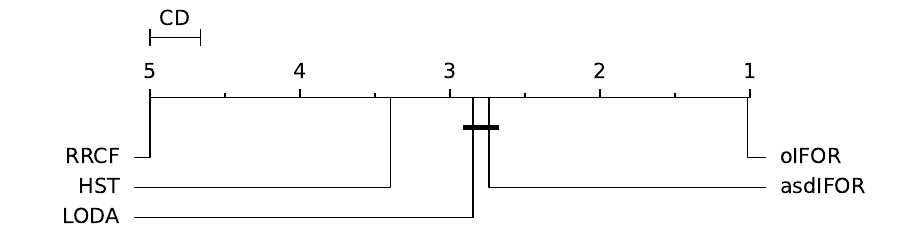}}
                    \caption{Times}
                    \label{fig:cd_times}
                    \end{center}
                \end{subfigure}
                \caption{Critical difference diagram for ROC AUCs and total execution times.}
                \label{fig:cd}
            \end{figure*}

            \subsubsection*{Anomaly detection in stationary data streams}
                \label{subsubsec:stationary}

                In~\cref{tab:performance} we show the median ROC AUC for each algorithm after processing each dataset, as well as the median total execution time for each algorithm to process the entire data stream at hand.
                The last two rows represent the median value and the mean rank among the total $330$ executions. We highlight the best row-wise result in bold.
                
                \oifor exhibits, by far, the lowest time complexity with respect to all the competitors at hand. In particular, when we compare \oifor to the second best method (\asdifor), we can notice that \oifor execution time is less than half in $4$ out of $5$ biggest datasets, and it is reduced by more than an order of magnitude in $4$ out of $6$ smallest ones. This result is confirmed by the critical difference diagram presented in~\cref{fig:cd_times}, which shows that \oifor is statistically better than all the others, while \loda and \asdifor are statistically equivalent. The statistical analysis has been conducted for $5$ populations (one for each algorithm) with $330$ paired total execution times. Critical difference diagrams are based on the post-hoc Nemenyi test, and differences between populations are significant when the difference of their mean rank is greater than the critical distance $CD = 0.336$. Mean ranks and critical diagrams have been generated via \textit{Autorank}~\cite{Herbold2020} library.
                In~\cref{fig:times} we sorted the median total execution times listed in~\cref{tab:performance} by dataset size, and we can appreciate the linear trend exhibited by all the algorithms.
    
                \cref{tab:performance} shows that \oifor, \asdifor and \rrcf exhibit the best detection performance over different datasets. While \oifor exhibits a slightly higher overall median ROC AUC, the Nemenyi test highlights that \asdifor is the most effective algorithm, and that \oifor and \rrcf are statistically equivalent (\cref{fig:cd_roc_aucs}).
                The exceptionally low performance of \hst on the Mulcross dataset in~\cref{tab:performance} is due to the fact that the size of anomaly clusters is large and that anomaly clusters have an equal or higher density compared to genuine ones in that dataset. This scenario, combined with the high default maximum depth value $\delta$, makes this situation particularly difficult for \hst to handle, as it is based on the opposite assumptions.
                
                \paragraph{Learning in the early stages of a data stream} In addition to the conventional evaluation of online anomaly detection algorithms based on their final anomaly detection capability, we highlight the initial learning speed demonstrated by various methods. We focused to the early stages of the data stream, aiming to comprehend how the various methods learn in a critical phase such as the initial one, and analyzed their performance within the first $1000$ samples of the stream.
                Solid curves in~\cref{fig:roc_auc_initial} show the median ROC AUC of each algorithm over all the $30$ executions and the $11$ datasets, for a total of $330$ runs. We computed the ROC AUCs at each time instant $\dot{t}$ using the scores from $t=1$ to $t=\dot{t}$. All the algorithms show a fast learning speed, since within the first $1000$ samples all of them get very close to the final median performance showed in~\cref{tab:performance}.
                The corresponding efficiency is shown in~\cref{fig:times_initial}, where all the algorithms exhibit similar trends, with \oifor being the fastest to adapt.

            \subsubsection*{Anomaly detection in non-stationary data streams}
                \label{subsubsec:non_stationary}

                \begin{figure*}[t]
                    \centerline{\includegraphics[width=.45\linewidth]{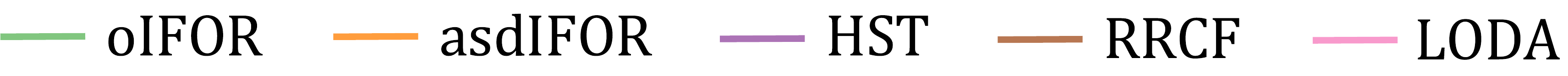}}
                    \begin{subfigure}[t]{\columnwidth}
                        \begin{center}
                        \centerline{\includegraphics[width=.8\columnwidth]{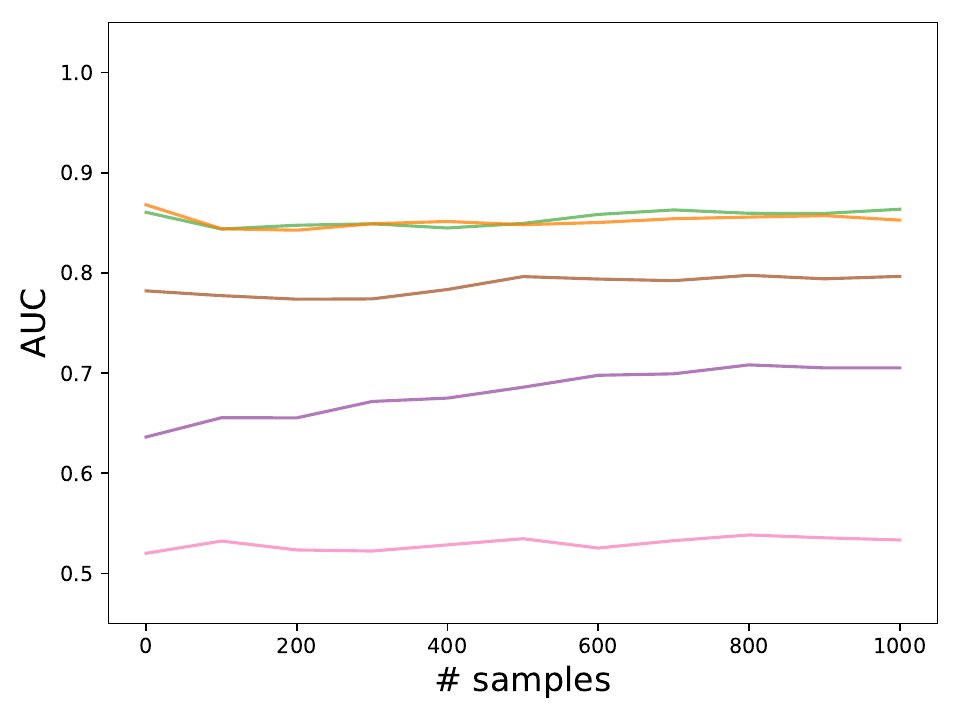}}
                        \caption{ROC AUCs}
                        \label{fig:roc_auc_initial}
                        \end{center}
                    \end{subfigure}
                    \hspace{-0.5cm}
                    \begin{subfigure}[t]{\columnwidth}
                        \begin{center}
                        \centerline{\includegraphics[width=.8\columnwidth]{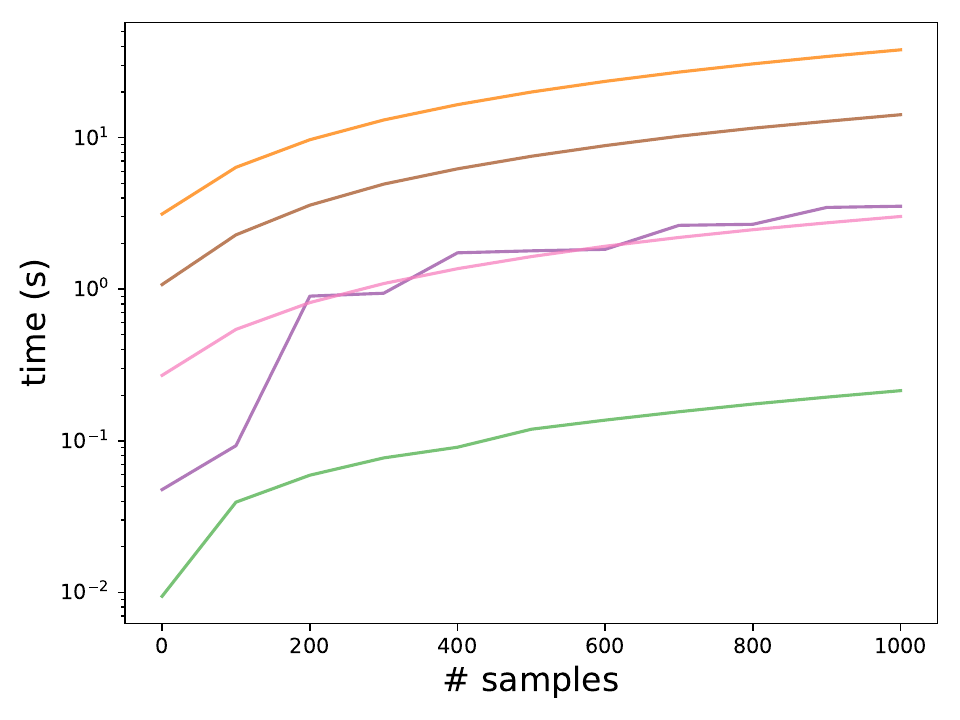}}
                        \caption{Times}
                        \label{fig:times_initial}
                        \end{center}
                    \end{subfigure}
                    \caption{Evolution of the median ROC AUCs and total execution times within the first $1000$ samples of the stream.}
                    \label{fig:performance_initial}
                \end{figure*}
                
                \begin{figure}[t]
                    \begin{center}
                    \centerline{\includegraphics[width=\columnwidth]{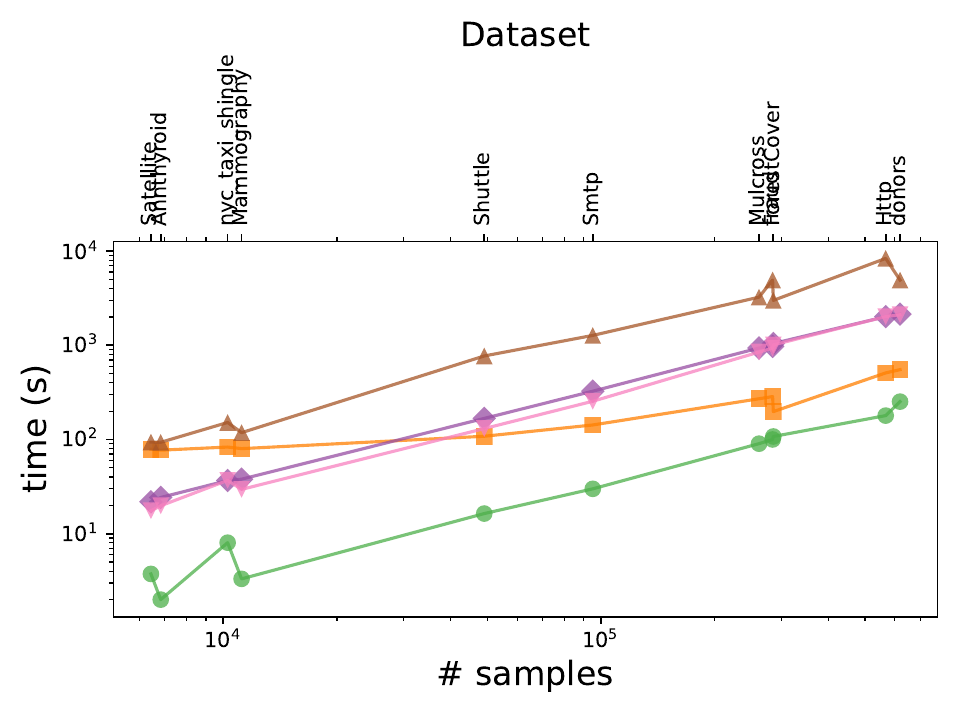}}
                    \centerline{\includegraphics[width=.9\linewidth, right]{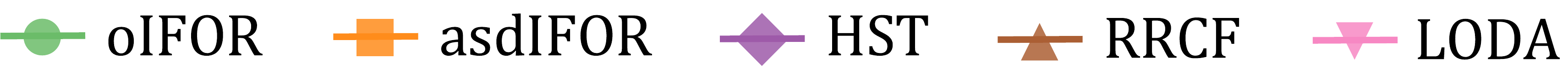}}
                    \caption{Total execution times ordered by dataset size.}
                    \label{fig:times}
                    \end{center}
                \end{figure}

                \begin{figure}[t]
                    \begin{center}
                    \centerline{\includegraphics[width=.9\linewidth, right]{pictures/experiments/real/legend.pdf}}
                    \centerline{\includegraphics[width=.9\linewidth]{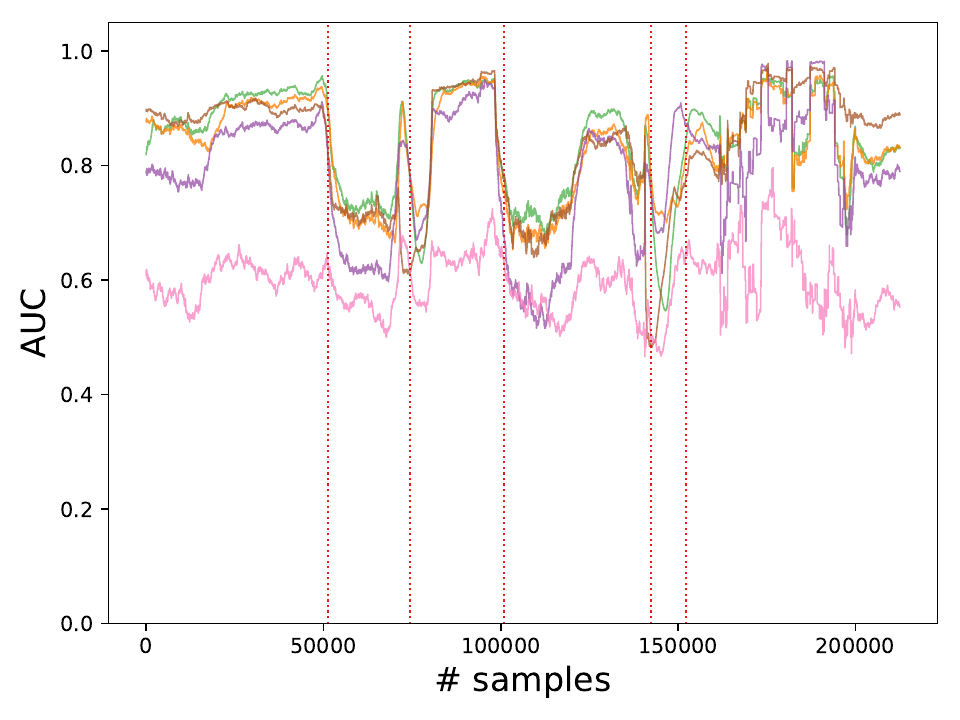}}
                    \caption{The performance of online anomaly detection methods is significantly influenced by changes in data distributions $\Phi_0$ and $\Phi_1$.}
                    \label{fig:convoluted_aucs}
                    \end{center}
                \end{figure}

                In~\cref{fig:convoluted_aucs} we show the median ROC AUC of each algorithm over $30$ executions on the INSECTS dataset. We are interested in investigating the instantaneous anomaly detection performance of all the algorithms, therefore we computed the ROC AUCs at each time instant $\dot{t}$ using the scores within a window of size $5000$ centered in $\dot{t}$, i.e., from $t=\dot{t}-2500$ to $t=\dot{t}+2500$. The choice of $5000$ for window size was made to guarantee that the resulting curves exhibit a satisfactory degree of smoothness.
                Vertical dotted lines represent the time instants when the change in distributions occur.
                \cref{fig:convoluted_aucs} shows that all the tested algorithms are affected in a similar way by sudden changes in the genuine and anomalous distributions $\Phi_0$ and $\Phi_1$, and we cannot identify a method that consistently maintains performance after a change. Although \loda is less affected by changes compared to the others, the overall low performance indicates that \loda struggles in learning the underlying distributions.
                The execution times of the algorithms are not influenced by distribution changes, and \onlineiforest remains the fastest option.

    \section{Conclusion and Future Works}
        \label{sec:conclusion_and_future_works}

        In this work we presented \onlineiforest, an anomaly detection algorithm specifically designed for the streaming scenario. \onlineiforest is an ensemble of histograms that dynamically adapt to the data distribution keeping only statistics about data points. Thanks to a sliding window, \onlineiforest is able to selectively forget old data points and update histograms accordingly.
        Extensive experiments showed that \onlineiforest features an extremely fast processing and learning speed while maintaining effectiveness comparable to that of state-of-the-art methods.
        The intuitive operational approach, coupled with its high speed, positions \onlineiforest as a good candidate for addressing real-world streaming anomaly detection challenges.

        As future work we aim to remove the sliding window $W$ while retaining the forgetting capabilities of \onlineiforest. Additionally, we seek to automate the selection of the number $\eta$ of points required to split histogram bins.

    \section*{Acknowledgements}
        This work was supported by the \textit{``PNRR-PE-AI FAIR"} and the \textit{``AI for Sustainable Port-city logistics (PNNR Grant P2022FLLPY)"} projects, both funded by the NextGeneration EU program.

    \section*{Impact Statement}
        This paper presents work whose goal is to advance the field of Machine Learning. There are many potential societal consequences of our work, none which we feel must be specifically highlighted here.
    
    \bibliography{references}
    \bibliographystyle{icml2024}
\end{document}